\documentclass[letterpaper]{article} 
\usepackage{aaai2026}  
\usepackage{times}  
\usepackage{helvet}  
\usepackage{courier}  
\usepackage[hyphens]{url}  
\usepackage{graphicx} 
\usepackage{multirow}
\usepackage{booktabs}
\usepackage[table]{xcolor}
\usepackage{amsmath}
\usepackage{amssymb}
\usepackage{natbib}  
\usepackage{caption} 
\usepackage{algorithm}
\usepackage{algorithmic}

\usepackage{newfloat}
\usepackage{listings}
\DeclareCaptionStyle{ruled}{labelfont=normalfont,labelsep=colon,strut=off} 
\floatstyle{ruled}
\newfloat{listing}{tb}{lst}{}
\floatname{listing}{Listing}
\title{MDiff4STR: Mask Diffusion Model for Scene Text Recognition}
\author {
    Yongkun Du\textsuperscript{\rm 1}\equalcontrib,
    Miaomiao Zhao\textsuperscript{\rm 2}\equalcontrib,
    Songlin Fan\textsuperscript{\rm 1,3},
    Zhineng Chen\textsuperscript{\rm 1}\thanks{Corresponding Author},
    Caiyan Jia\textsuperscript{\rm 2},
    Yu-Gang Jiang\textsuperscript{\rm 1}
}
\affiliations {
    \textsuperscript{\rm 1}Institute of Trustworthy Embodied AI, Fudan University, China\\
    \textsuperscript{\rm 2}School of Computer Science and Technology, Beijing Jiaotong University, China\\
    \textsuperscript{\rm 3}China Mobile Shanghai ICT Co., Ltd., China\\
    ykdu23@m.fudan.edu.cn, \{zhinchen,slfan,ygj\}@fudan.edu.cn, \{miaomiao\_zhao,cyjia\}@bjtu.edu.cn
}

\usepackage{bibentry}

\begin{document}

\maketitle

\begin{abstract}
Mask Diffusion Models (MDMs) have recently emerged as a promising alternative to auto-regressive models (ARMs) for vision-language tasks, owing to their flexible balance of efficiency and accuracy. In this paper, for the first time, we introduce MDMs into the Scene Text Recognition (STR) task. We show that vanilla MDM lags behind ARMs in terms of accuracy, although it improves recognition efficiency. To bridge this gap, we propose MDiff4STR, a Mask Diffusion model enhanced with two key improvement strategies tailored for STR. Specifically, we identify two key challenges in applying MDMs to STR: noising gap between training and inference, and overconfident predictions during inference. Both significantly hinder the performance of MDMs. To mitigate the first issue, we develop six noising strategies that better align training with inference behavior. For the second, we propose a token-replacement noise mechanism that provides a non-mask noise type, encouraging the model to reconsider and revise overly confident but incorrect predictions. We conduct extensive evaluations of MDiff4STR on both standard and challenging STR benchmarks, covering diverse scenarios including irregular, artistic, occluded, and Chinese text, as well as whether the use of pretraining. Across these settings, MDiff4STR consistently outperforms popular STR models, surpassing state-of-the-art ARMs in accuracy, while maintaining fast inference with only three denoising steps.

\end{abstract}

\begin{links}
    \link{Code}{https://github.com/Topdu/OpenOCR}
\end{links}

\section{Introduction}

\begin{figure}[t]
  \centering
\includegraphics[width=0.45\textwidth]{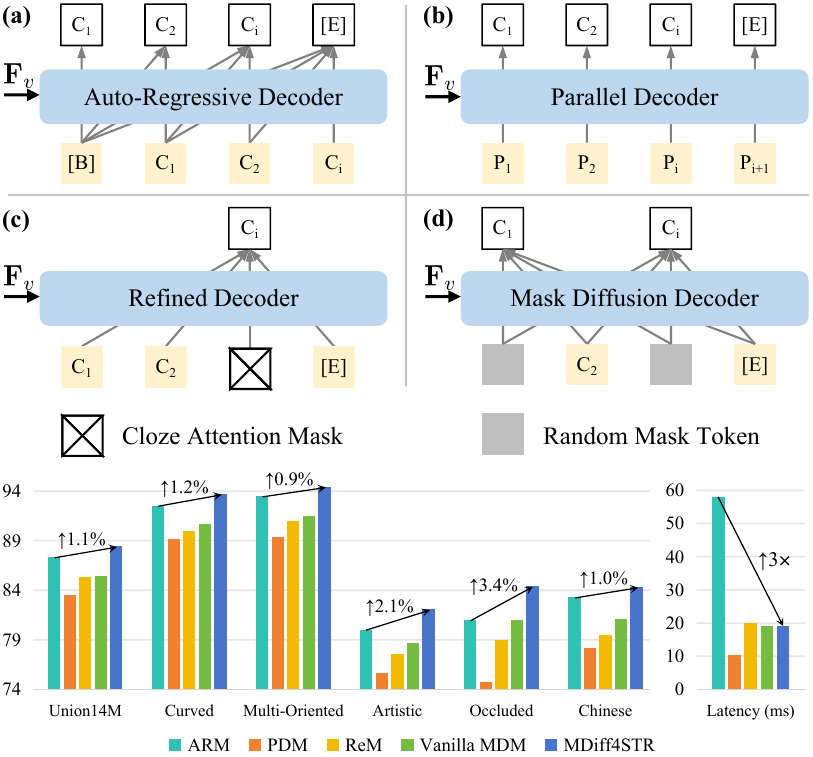} 
\caption{(a) Auto-regressive models (ARMs), (b) Parallel decoding models (PDMs), (c) BERT-like refinement models (ReMs), (d) Mask diffusion models (MDMs). $\mathbf{F}_v$ means visual features. MDMs learn to reconstruct character sequences from partially masked inputs through a denoising process,  capturing more flexible and comprehensive omnidirectional dependencies than ARMs and refinement models.}
\label{fig:fig1}
\end{figure}

Scene Text Recognition (STR), as a base task in Optical Character Recognition (OCR) systems, has remained a focal point of research in computer vision. In natural scenes, STR faces a wide range of complex challenges ~\cite{ChenJZLW21str_survey}, including curved and distorted text, varying orientations, occlusions, image blur, and artistic fonts. To address these issues, researchers have proposed a wealth of innovative solutions that significantly enhance the robustness and accuracy of recognition models in real-world applications. 

Among STR methods, auto-regressive models (ARMs) ~\cite{shi2019aster,Sheng2019nrtr,li2019sar,yue2020robustscanner,jiang2023revisiting,xie2022toward_cornertrans,zheng2023tps++,zheng2024cdistnet,Xu_2024_CVPR_OTE,yang2024class_cam,zhou2024cff,du2024igtr,du2024smtr} have emerged as one of the most prominent due to their strong sequence modeling capabilities and achieved state-of-the-art results across standard and challenging benchmarks \cite{Wang_2021_visionlan,jiang2023revisiting,chen2021benchmarking}. However, the inherently sequential nature of ARMs limits their decoding efficiency.

Recently, mask diffusion models (MDMs)~\cite{ShiHWDT24_simmaskdiff_dis,SahooASGMCRK24_simmaskdiff,llada,lladav}, a novel non-auto-regressive paradigm, have demonstrated superior performance in both efficiency and accuracy. On the one hand, as shown in Fig.~\ref{fig:fig1}(d), MDMs learn to reconstruct original sequences from partially masked inputs by progressively denoising them. This formulation not only overcomes the unidirectional (left-to-right) modeling limitation of ARMs, but also surpasses the rigid bidirectional modeling of BERT-like models, enabling the capture of more flexible and comprehensive omnidirectional dependencies ~\cite{ShiHWDT24_simmaskdiff_dis,SahooASGMCRK24_simmaskdiff}. Given that STR is fundamentally reliant on strong language understanding, MDMs show great potential in offering a novel and promising paradigm for enhancing STR performance. On the other hand, its denoising process is highly efficient and controllable, allowing accurate predictions to be produced within only a few steps.

In this paper, we introduce, for the first time, MDMs into the STR task. However, experimental results reveal that the vanilla MDM offers notable advantages in inference efficiency, but it still falls short in recognition accuracy compared to ARMs. To uncover the causes of this performance gap, we identify two key limitations: (1) Noising gap between training and inference. MDMs are typically trained with randomly noisy, while during inference, the model is exposed to structured and deterministic noisy patterns. These patterns are rarely encountered during training, resulting in poor generalization and degraded recognition performance; (2) Overconfident predictions during inference. We observe that MDMs tend to assign excessively high confidence scores to their predictions even when incorrect. This overconfidence hampers the effectiveness of the confidence-based remask mechanism, making it difficult for the model to identify and correct earlier mistakes, thereby undermining the overall efficacy of the multi-step denoising process.

To address the aforementioned challenges, we propose MDiff4STR, a Mask Diffusion Model tailored for STR. It is a novel MDM paradigm for STR incorporating two key innovations. First, we introduce six noising strategies for training. They accurately simulate the noising patterns encountered during inference, effectively mitigating the noising gap. Second, we introduce a token-replacement noise mechanism, a novel noise type distinct from masking. This mechanism enables the model to reconsider and correct its own overconfident yet incorrect predictions, leading to more accurate recognition results.

Experiments on multiple public STR benchmarks demonstrate that MDiff4STR consistently outperforms popular STR models, surpassing state-of-the-art ARMs in accuracy. Meanwhile, it also maintains fast inference because it only
requires three denoising steps. These results indicate that MDiff4STR establishs a novel STR paradigm, simultaneously achieving superior accuracy and high efficiency. The contributions of this paper are threefold:
\begin{itemize}
    \item To the best of our knowledge, we introduce MDMs into the STR task for the first time, pioneering a novel alternative to ARMs that enables a flexible trade-off between accuracy and efficiency.
    \item We identify two key challenges in applying the the MDM in STR: noising gap between training and inference, and overconfident predictions during inference. To address these issues, we propose six tailored noise strategies as well as a token-replacement noise mechanism.
    \item We present MDiff4STR, a MDM-based framework for STR that outperforms state-of-the-art ARMs in recognition accuracy while achieving 3$\times$ faster inference, establishing a novel paradigm for the task.
\end{itemize}

\section{Related Work}

Scene Text Recognition (STR), as a typical vision-language task, often heavily relies on linguistic context to achieve accurate recognition. Most efforts~\cite{shi2016rare,shi2019aster,Sheng2019nrtr,li2019sar,yue2020robustscanner,jiang2023revisiting,xie2022toward_cornertrans,zheng2024cdistnet,Xu_2024_CVPR_OTE,yang2024class_cam,zhou2024cff,du2024igtr,du2024smtr,su2025lranetpp} integrate language modeling capabilities into STR using autoregressive models (ARMs). As illustrated in Fig.~\ref{fig:fig1}(a), ARMs predict characters through iterative decoding, explicitly modeling the contextual dependencies between characters. However, the inherently sequential nature of ARMs limits their decoding efficiency. To overcome this limitation and improve inference speed, researchers have proposed parallel decoding models (PDMs) ~\cite{wang2020aaai_dan,Wang_2021_visionlan,mgpstr,du2023cppd,ijcai2023LPV,qiao2021pimnet,ipad}, as illustrated in Fig.~\ref{fig:fig1}(b). These models abandon token-to-token dependency modeling and instead generate the entire character sequence simultaneously, resulting in a significant boost in decoding speed. However, due to the absence of contextual modeling, their recognition accuracy generally lags behind that of ARMs. To strike a better balance between speed and accuracy, some works \cite{yu2020srn,fang2021abinet,BautistaA22PARSeq,MATRN,Wei_2024_busnet} have introduced BERT-like architectures \cite{bert}, as depicted in Fig.~\ref{fig:fig1}(c). These models first produce an initial results in parallel and then refine the predictions by incorporating contextual information. While this strategy mitigates the lack of contextual understanding in purely parallel models, it can be sensitive to initial prediction errors, which may propagate during refinement ~\cite{jiang2023revisiting,du2024svtrv2}, limiting their ability to outperform ARMs. MDM, distinct from them, learns to reconstruct character sequences from partially masked inputs through a denoising process, capturing more flexible and comprehensive omnidirectional dependencies than ARMs and refinement models. Moreover, its denoising process is highly efficient and controllable, enabling accurate predictions to be produced in just a few steps, thereby achieving faster inference.

\begin{figure}[t]
  \centering
\includegraphics[width=0.48\textwidth]{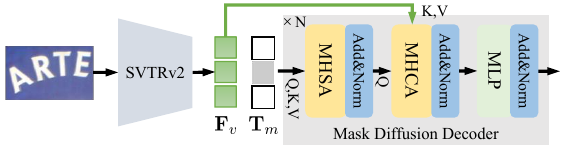} 
  \caption{The network of MDiff4STR. $\mathbf{F}_v$ and $\mathbf{T}_m$ denote the visual features and the noised tokens, respectively.}
  \label{fig:network}
\end{figure}

\begin{figure*}[t]
  \centering
\includegraphics[width=0.94\textwidth]{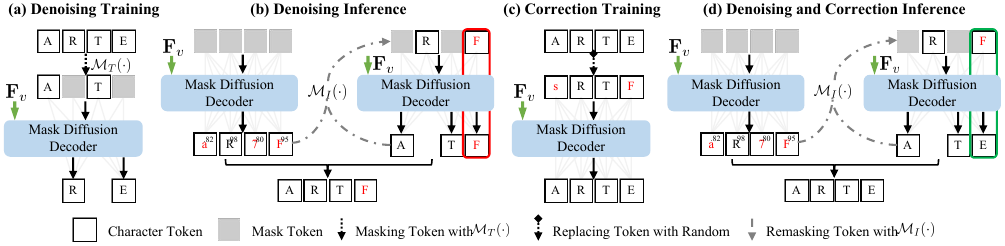} 
  \caption{(a) and (b) denote the denoising training and inference of the vanilla MDM, respectively. (c) depicts the error-correction training enabled by our proposed token-replacement noise mechanism. MDiff4STR jointly leverages (a) denoising and (c) error-correction training to achieve (d) the denoising process augmented with corrective capability. Red boxes indicate errors caused by overconfident predictions, whereas green boxes highlight correct reasoning performed by MDiff4STR. $\mathcal{M}_T$ and $\mathcal{M}_I$ denote the mask strategy for training and the remask strategy for inference, respectively.}
  \label{fig:MDiff4STR_overview}
\end{figure*}

\section{Method}

Our MDiff4STR is a novel STR method based on the recent MDM. Fig.~\ref{fig:network} provides details of the model architecture, while Fig.~\ref{fig:MDiff4STR_overview} illustrates the overall training and inference pipeline. Given an input text image $\mathbf{X} \in \mathbb{R}^{H \times W \times 3}$ and its corresponding text label $\mathbf{Y} = \{y_1,y_2,\dots,y_L\} \in \mathbb{V}^{L}$, where $\mathbb{V}$ denotes the vocabulary, we uses SVTRv2~\cite{du2024svtrv2}, a visual encoder specially designed for STR, to extract image features  $\mathbf{F}_v \in \mathbb{R}^{\frac{H}{8} \times \frac{W}{4} \times D}$. On the textual side, the character sequence $\mathbf{Y}$ is noised by the mask token to produce noised token sequence $\mathbf{T}_m \in \mathbb{R}^{L \times D}$. The mask diffusion decoder then performs a denoising process conditioned on the visual features to predict the final recognition.

\subsection{Vanilla Mask Diffusion Model}

In the denoising training of the vanilla MDM, the character sequence $\mathbf{Y}$ is partially and randomly masked with a mask token $\texttt{[MASK]}$, resulting in a noised version $\mathbf{Y}_m = \mathcal{M}_T(\mathbf{Y}) \in \mathbb{V}^{L}$. $\mathcal{M}_T$ means the mask strategy for training. The mask diffusion decoder then recovers the original sequence $\mathbf{Y}$ from the noised sequence $\mathbf{Y}_m$, conditioned on the image features $\mathbf{F}_v$. The entire training process can be formally described as:
\begin{gather*}
    \mathbf{F}_v = \text{SVTRv2}(\mathbf{X}),~\mathbf{T}_m = \text{Embedding}(\mathbf{Y}_m) \\
    \mathbf{\tilde{T}} = \text{MDiffDecoder}(\mathbf{F}_v, \mathbf{T}_m),~\mathbf{\tilde{Y}} = \text{Classifier}(\mathbf{\tilde{T}})
\end{gather*}
where $\text{Embedding}(\cdot)$ is a learnable character embedding layer that maps characters into a vector space, and $\text{Classifier}(\cdot)$ maps the decoded tokens back to character.

The inference process is modeled as $K$-step denoising diffusion process. The first step, the token sequence is entirely set to mask tokens, representing a completely unknown character sequence. Then, $K-1$ remask steps are applied to progressively generate the final prediction $\mathbf{\tilde{Y}}^K$:
\begin{gather*}
\mathbf{Y}_m^1 = \texttt{[MASK]}^{\otimes L},~ \mathbf{T}_m^i = \text{Embedding}(\mathbf{Y}_m^i) \\ 
\mathbf{\tilde{T}}^i = \text{MDiffDecoder}(\mathbf{F}_v, \mathbf{T}_m^i),~ \mathbf{\tilde{Y}}^i = \text{Classifier}(\mathbf{\tilde{T}}^i) \\
\mathbf{Y}_m^{i+1} = \mathcal{M}_I(\mathbf{\tilde{Y}}^i)
\end{gather*}

Depending on the remask strategy for inference $\mathcal{M}_I$, the MDM can flexibly implement multiple decoding paradigms. Specifically, as shown in Fig.~\ref{fig:mask7}(b), when the full mask is adopted, parallel decoding (MDiff-PD) is formed. Fig.~\ref{fig:mask7}(c/d) respectively illustrate the forward and backward auto-regressive denoising processes. To maintain consistency with previous ARMs, we only consider the denoising behaviour in Fig.~\ref{fig:mask7}(c) as auto-regressive decoding (MDiff-AR). Fig.~\ref{fig:mask7}(e) corresponds to BERT-like refinement decoding (MDiff-Re).
In addition, the MDM introduces a unique and efficient decoding method: confidence-guided remask. Fig.~\ref{fig:mask7}(f) demonstrates low-confidence remask (MDiff-LC), where after each denoising step, low-confidence tokens denoting below the average confidence score are remasked based on predicted confidence and fed into the next iteration. This approach leverages the MDM's ability to repair uncertain predictions.
However, this strategy is prone to the confidence trap, where certain tokens are repeatedly remasked in each iteration, leading to generation stagnation. To avoid this issue, MDiff-BLC (Fig.~\ref{fig:mask7}(g)) adopts remask for low-confidence tokens within a fixed-size local block, effectively avoiding the confidence trap. Here, the block size is set to $\frac{L}{K}$.

\subsection{Training Mask Strategies ($\mathcal{M}_T$)} 

\begin{figure}[t]
  \centering
\includegraphics[width=0.48\textwidth]{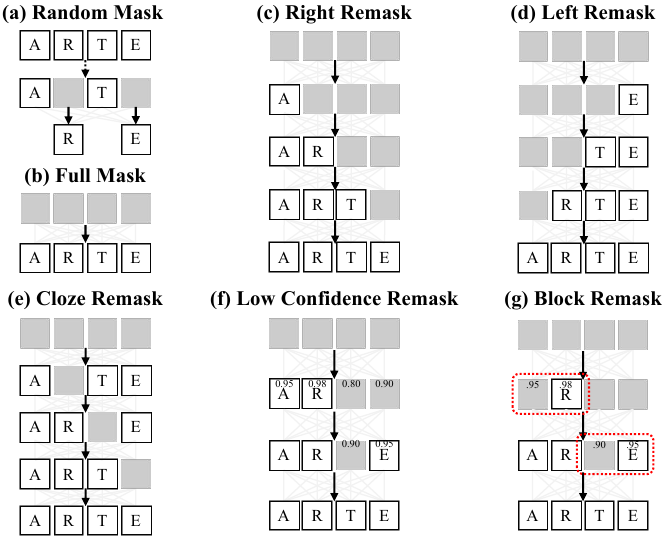} 
  \caption{(a) illustrates random token masking for training. (b) shows the full mask strategy, which is used as the initial denoising step during inference. Subfigures (b–g) present various remasking strategies for inference and also server as noise strategies in training to eliminate the noising gap.}
  \label{fig:mask7}
\end{figure}

During training, as shown in Fig.~\ref{fig:mask7}(a), the vanilla MDM typically adopts random mask strategies to corrupt input sequences. In contrast, the inference process begins with a fully masked token sequence (Fig.~\ref{fig:mask7}(b)) and then is followed by a multi-step denoising process using the remask strategies $\mathcal{M}_I$ in Fig.~\ref{fig:mask7}(c/d/e/f/g).

The full mask and remask strategies for inference are included in the training of the randomized mask strategy, but with minimal frequency. This creates noising gap between training and inference, leading to poor generalization. To address this issue, we use seven mask strategies (noted as $\mathcal{M}_T$)), i.e. Fig.~\ref{fig:mask7}(a/b/c/d/e/f/g), from which one is uniformly sampled during training for each input sequence. This design significantly improves the model’s robustness and adaptability to the remask patterns encountered during inference.

\subsection{Token-Replacement Noise Mechanism}

MDMs tend to be overconfident and often assign high confidence score to incorrect predictions. For example, the character ``F'' in Fig.~\ref{fig:MDiff4STR_overview}(b) is predicted with a high confidence score of 0.95, preventing it from being remasked. As a result, the incorrect prediction cannot be corrected in subsequent steps, ultimately leading to a misrecognized output.

To address this issue, we propose a token-replacement noise mechanism. Specifically, we randomly replace certain characters in the original sequence $\mathbf{Y}$ with another character to construct a corrupted sequence $\mathbf{Y}_r \in \mathbb{V}^{L}$. This sequence is then embedded into token representations $\mathbf{T}_r = \text{Embedding}(\mathbf{Y}_r) \in \mathbb{R}^{L \times D}$, which serve as input to simulate erroneous predictions. An example of such corruption is illustrated in Fig.~\ref{fig:MDiff4STR_overview}(c), demonstrating how the model handles incorrect tokens during training. This method closely emulates the ``erroneous yet high-confidence" scenarios encountered during inference. Subsequently, the model is trained to correct these incorrect tokens as part of the denoising task:
$$
\mathbf{\tilde{T}} = \text{MDiffDecoder}(\mathbf{F}_v, \mathbf{T}_r),~ \mathbf{\tilde{Y}} = \text{Classifier}(\mathbf{\tilde{T}})
$$

As shown in Fig.~\ref{fig:MDiff4STR_overview}(d), our token-replacement noise mechanism enables robust error correction during iterative decoding. Consequently, even if the error token ``F'' is not remasked in a given iteration, the model can still rectify it in subsequent iterations. Additionally, the token-replacement noise mechanism presents a novel and effective noise type distinct from simple masking, suggesting that MDM noise paradigms can extend beyond masking to unlock new research avenues.

\subsection{Training Objectives}

MDiff4STR integrates two distinct training objectives, corresponding to the structures shown in Fig.~\ref{fig:MDiff4STR_overview}(a) and \ref{fig:MDiff4STR_overview}(c), respectively. They are defined as follows:

\begin{align*}
\mathcal{L}_{\text{denoising}} &= -\frac{1}{l_1}\sum\nolimits_{i=1}^{L} \mathbf{1}[\mathbf{Y}_{l_1}^{i} = \mathbf{M}] \log p_{\theta}(\mathbf{Y}^i \mid \mathbf{Y}_{l_1}) \\
\mathcal{L}_{\text{correction}} &= -\frac{1}{L}\sum\nolimits_{i=1}^{L} \log p_{\theta}(\mathbf{Y}^i \mid \mathbf{Y}_{l_2}) \\
\mathcal{L}_{\text{total}} &= \mathcal{L}_{\text{denoising}} + \mathcal{L}_{\text{correction}}
\end{align*}

Here, $\mathbf{Y}_{l_1} = \mathbf{Y}_m$ and $\mathbf{Y}_{l_2} =\mathbf{Y}_r$. The terms $l_1$ and $l_2$ denote the number of tokens that are masked and the number of tokens that are randomly replaced with other characters, respectively. Both $l_1$ and $l_2$ are sampled uniformly from the range $[0, L]$, where $L$ is the length of the character sequence.

In the denoising loss $\mathcal{L}_{\text{denoise}}$, $\mathbf{1}[\mathbf{Y}_{l_1}^{i} = \mathbf{M}]$ is an indicator function that equals 1 only if the token at position $i$ is a \texttt{[MASK]} token. This ensures that the model is supervised exclusively on the masked positions. The term $p_{\theta}(\mathbf{Y}^i \mid \mathbf{Y}_{l_1})$ represents the probability of the model predicting the original token $\mathbf{Y}^i$ at position $i$, given the masked input sequence $\mathbf{Y}_{l_1}$. Conversely, for the correction loss $\mathcal{L}_{\text{correction}}$, the supervision is applied across the entire sequence. This is critical because, during inference, the model has no prior knowledge of which tokens are incorrect. Therefore, the correction training requires the model to make predictions for all tokens, compelling it to learn error correction capabilities under conditions of unknown perturbations. The final training objective $\mathcal{L}_{\text{total}}$ is the sum of these two losses.

\begin{figure*}[t]
  \centering
\includegraphics[width=0.99\textwidth]{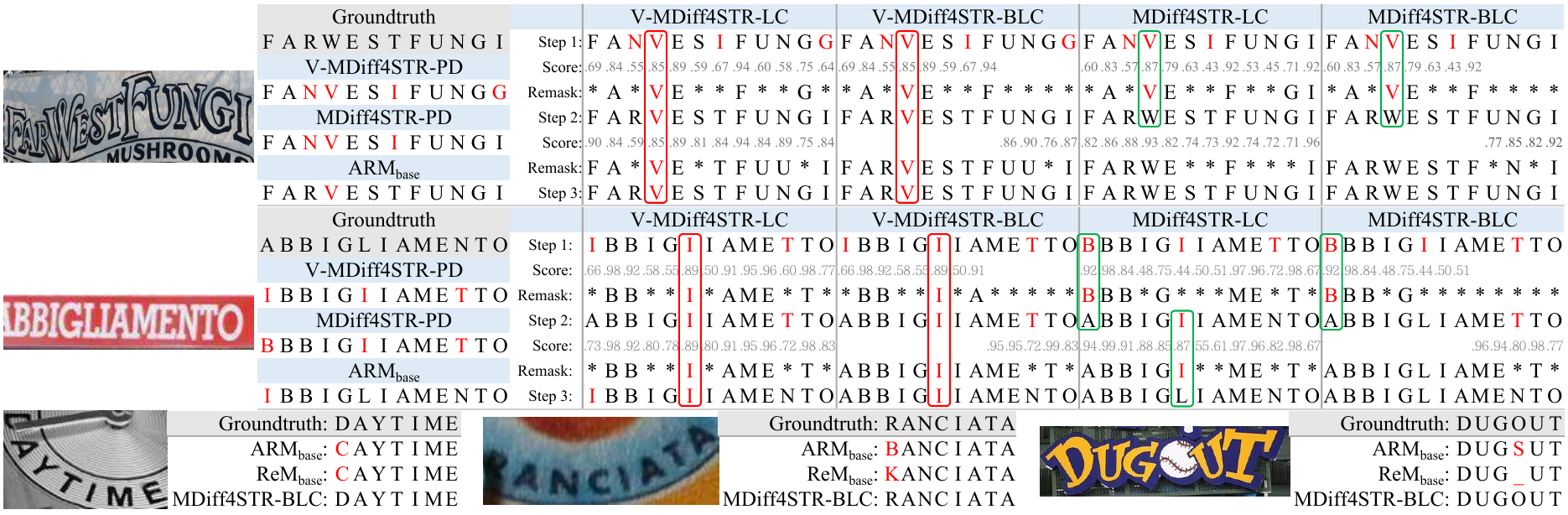} 
  \caption{The first two figures present the denoising process of MDiff4STR, while the last three demonstrate its reasoning advantage over ARM and ReM in omnidirectional contextual modeling involving occluded or artistic text recognition. V-MDiff4STR represents indicates that the token-replacement noise mechanism is not used during training. Red characters and underline denote the misrecognition and misspelling, respectively. Red boxes indicate errors caused by overconfident predictions, whereas green boxes highlight correct reasoning performed by MDiff4STR. * indicates tokens that are remasked as the \texttt{[mask]}.}
  \label{fig:case}
\end{figure*}

\section{Experiments}

\subsection{Datasets and Implementation Details}
\label{sec:Implementation}

For English recognition, we train MDiff4STR on \textit{U14M-Filter} \cite{du2024svtrv2}, which without data leakage. Then, we evaluate MDiff4STR across multiple benchmarks covering diverse scenarios. They are: 1) six common regular and irregular benchmarks (\textit{Com}), including ICDAR 2013 (\textit{IC13})~\cite{icdar2013}, Street View Text (\textit{SVT})~\cite{Wang2011SVT}, IIIT5K-Words (\textit{IIIT5K})~\cite{IIIT5K}, ICDAR 2015 (\textit{IC15})~\cite{icdar2015}, Street View Text-Perspective (\textit{SVTP})~\cite{SVTP} and \textit{CUTE80}~\cite{Risnumawan2014cute}. For IC13 and IC15, we use the versions with 857 and 1811 images, respectively; 2) Union14M-Benchmark (\textit{U14M})~\cite{jiang2023revisiting}, which includes seven challenging subsets: \textit{Curve}, \textit{Multi-Oriented}, \textit{Artistic}, \textit{Contextless}, \textit{Salient}, \textit{Multi-Words} and \textit{General}; 3) occluded scene text dataset (\textit{OST})~\cite{Wang_2021_visionlan}, which requires contextual reasoning for accurate recognition.

For Chinese recognition, we use \textit{BCTR}~\cite{chen2021benchmarking} including four scenarios. We trained the model on integration of the four subsets and then evaluated it on the test sets: \textit{Scene}, \textit{Web}, \textit{Document} (\textit{Doc}) and \textit{Hand-Writing} (\textit{HW}).

We use AdamW optimizer~\cite{adamw} with a weight decay of 0.05 for training. The learning rate (LR) is set to $5\times 10^{-4}$ and batchsize is set to 1024. One cycle LR scheduler~\cite{cosine} with 1.5/4.5 epochs linear warm-up is used in all the 40/100 epochs for English and Chinese model, respectively. Word accuracy is used as the evaluation metric. Data augmentation like rotation, perspective distortion, motion blur, and gaussian noise, are randomly performed. The maximum text length is set to 25. The vocabulary size $|\mathbb{V}|$ is set to 94 for English and 6624~\cite{ppocrv3} for Chinese. All models are trained on 4 RTX 3090 GPUs.

\subsection{Ablation Study}

Tab.~\ref{tab:K_N} presents the impact of the number of denoising steps ($K$) and decoder layers ($N$) on both accuracy and inference speed. Based on the results, we set $K = 3$ and $N = 6$ to achieve a balance between accuracy and efficiency.

To fairly evaluate the effectiveness of MDiff4STR, we implemented three baseline models: ARM$_{base}$,  PDM$_{base}$, and ReM$_{base}$, using the exact same model architecture shown in Fig.~\ref{fig:network}(a) and training setting present in Sec.~\ref{sec:Implementation}. As shown in the first part of Tab.~\ref{tab:mask_reflect}, ARM$_{base}$ achieves superior performance compared to all previous methods (refer to Tab.~\ref{tab:sota}), which supports the robustness of our baseline models.

\begin{table}[t]\footnotesize
\centering
\setlength{\tabcolsep}{4.5pt}{
\begin{tabular}{c|c|cc|cc|cc|c}
\toprule
\multicolumn{2}{c|}{\multirow{2}{*}{}} & \multicolumn{2}{c|}{\textit{Com}} & \multicolumn{2}{c|}{\textit{U14M}} & \multicolumn{2}{c|}{\textit{OST}} & \multirow{2}{*}{\textit{Time}} \\
\multicolumn{2}{c|}{}                  & LC         & BLC        & LC          & BLC        & LC         & BLC        &                       \\
\hline
\multirow{8}{*}{\textit{K}}         & 1        & 96.88      & 96.88      & 86.69       & 86.69      & 81.31      & 81.31      &     10.52                  \\
& 2        & 97.19      & 97.19      & 88.05       & 88.05      & 83.69      & 83.69      &  15.56                     \\
& \cellcolor[HTML]{D9E1F4}3        & \cellcolor[HTML]{D9E1F4}97.29      & \cellcolor[HTML]{D9E1F4}97.30      & \cellcolor[HTML]{D9E1F4}88.37       & \cellcolor[HTML]{D9E1F4}88.44      & \cellcolor[HTML]{D9E1F4}84.21      & \cellcolor[HTML]{D9E1F4}84.25      &  \cellcolor[HTML]{D9E1F4}19.21                     \\
& 4        & 97.31      & 97.28      & 88.42       & 88.59      & 84.19      & 84.27      &    23.11                   \\
& 5        & 97.31      & 97.28      & 88.42       & 88.50      & 84.19      & 84.42      &  25.70                     \\
& 6        & 97.31      & 97.29      & 88.42       & 88.57      & 84.19      & 84.33      &     28.74                  \\
& 7        & 97.31      & 97.24      & 88.42       & 88.63      & 84.19      & 84.09      &     30.64                  \\
& 8        & 97.31      & 97.24      & 88.42       & 88.65      & 84.19      & 84.11      &   32.74                    \\
\hline
\multirow{4}{*}{\textit{N}}       
& 2        &  96.73          &     96.71       &     87.92        &  87.95          &  81.66          &     81.64       &   12.26                    \\
& 4        &  96.92          &   97.07         &   88.04          &  88.10          &    82.30        & 82.33           &  16.07                     \\
& \cellcolor[HTML]{D9E1F4}6        & \cellcolor[HTML]{D9E1F4}97.29      & \cellcolor[HTML]{D9E1F4}97.30      & \cellcolor[HTML]{D9E1F4}88.37       & \cellcolor[HTML]{D9E1F4}88.44      & \cellcolor[HTML]{D9E1F4}84.21      & \cellcolor[HTML]{D9E1F4}84.25      & \cellcolor[HTML]{D9E1F4}19.21                      \\
& 8        &      97.00      &  97.00          &    87.63         &  87.80           &   82.65         &    82.72        &  23.04                    \\
\bottomrule
\end{tabular}}
\caption{Ablation study on the number of denoising steps $K$ and decoder layers $N$ in MDiff4STR.}
\label{tab:K_N}
\end{table}

\begin{table}[H]\footnotesize
\centering
\setlength{\tabcolsep}{1.0pt}{
\begin{tabular}{l|cc|ccc|c}
\toprule
Decoding  & $\mathcal{M}_T$         & TRN & \textit{Com}         & \textit{U14M}        & \textit{OST} & \textit{Time}\\
\hline
PDM$_{base}$       &    -        &   -      & 95.78 & 83.54 & 74.77 &  10.52    \\
ARM$_{base}$       &    -        &   -      & 96.88 & 87.34 & 81.03 &  57.95    \\
ReM$_{base}$       &   -         &   -      & 96.05 & 84.91 & 78.98 &  20.11    \\
\hline
MDiff-PD  & \textit{R}     & ×       & 95.76 & 84.36 & 76.70 &  10.52    \\
MDiff-AR  & \textit{R}     & ×       & 96.31 & 85.62 & 77.79 &  66.35    \\
MDiff-Re  & \textit{R}     & ×       & 96.24 & 84.93 & 79.41 &  20.11    \\
MDiff-LC  & \textit{R}     & ×       & 96.43 & 85.33 & 79.41 &  19.21    \\
MDiff-BLC & \textit{R}     & ×       & 96.42 & 85.42 & 79.93 &  19.21    \\
\hline
MDiff-PD  & \textit{R+All} & ×       & 96.03$_{\uparrow0.27}$ & 86.00$_{\uparrow1.64}$ & 78.93$_{\uparrow2.23}$ &   10.52   \\
MDiff-AR  & \textit{R+All} & ×       & 96.58$_{\uparrow0.26}$ & 87.13$_{\uparrow1.51}$ & 79.94$_{\uparrow2.14}$ &   66.35   \\
MDiff-Re  & \textit{R+All} & ×       & 96.77$_{\uparrow0.53}$ & 86.84$_{\uparrow1.91}$ & 81.45$_{\uparrow2.04}$ &  20.11    \\
MDiff-LC  & \textit{R+All} & ×       & 96.95$_{\uparrow0.53}$ & 86.98$_{\uparrow1.65}$ & 81.86$_{\uparrow2.45}$ &  19.21    \\
MDiff-BLC & \textit{R+All} & ×       & 96.98$_{\uparrow0.57}$ & 87.09$_{\uparrow1.67}$ & 81.92$_{\uparrow2.00}$ &  19.21    \\

\hline
MDiff-PD  & \textit{R+All} & \checkmark       & 96.88$_{\uparrow1.13}$ & 86.69$_{\uparrow2.33}$ & 81.31$_{\uparrow4.61}$ &  10.52    \\
MDiff-AR  & \textit{R+All} & \checkmark       & 97.09$_{\uparrow0.78}$ & \textbf{88.63}$_{\uparrow3.01}$ & 82.47$_{\uparrow4.68}$ &  66.35    \\
MDiff-Re  & \textit{R+All} & \checkmark       & 97.12$_{\uparrow0.88}$ & 88.16$_{\uparrow3.23}$ & 83.90$_{\uparrow4.49}$ &  20.11    \\
MDiff-LC  & \textit{R+All} & \checkmark       & 97.29$_{\uparrow0.87}$ & 88.37$_{\uparrow3.04}$ & 84.21$_{\uparrow4.80}$ &  19.21    \\
MDiff-BLC & \textit{R+All} & \checkmark       & \textbf{97.30}$_{\uparrow0.88}$ & 88.44$_{\uparrow3.02}$ & \textbf{84.25}$_{\uparrow4.33}$ & 19.21    \\
\multicolumn{7}{c}{\setlength{\tabcolsep}{4.0pt}{\begin{tabular}{c|ccccccc}
\hline
&Base&+(b)&+(c)&+(d)&+(e)&+(f)&+(g) \\
\midrule
U14M&85.42&$\uparrow$1.04&$\uparrow$1.37&$\uparrow$1.41&$\uparrow$1.46&$\uparrow$1.53&$\uparrow$1.67
\end{tabular}}} \\
\bottomrule
\end{tabular}}
\caption{\textbf{Top:} Influence of the training mask strategies $\mathcal{M}_T$ and the token-replacement noise mechanism (TRN). \textit{R} denotes the random mask strategy, and \textit{R+All} denotes using all of the mask strategies in Fig.~\ref{fig:mask7} for training. \textbf{Bottom:} Ablation on the six masks in Fig.~\ref{fig:mask7}. Here, +(*) means using the mask in Fig.~\ref{fig:mask7}(*) gradually.}
\label{tab:mask_reflect}
\end{table}

\begin{table*}[t]\footnotesize
\centering
\setlength{\tabcolsep}{2.8pt}{
\begin{tabular}{c|ll|ccccccc|cccccccc|c|c}
\multicolumn{20}{c}{\setlength{\tabcolsep}{2.3pt}{\begin{tabular}{
>{\columncolor[HTML]{FFFFC7}}c 
>{\columncolor[HTML]{FFFFC7}}c 
>{\columncolor[HTML]{FFFFC7}}c 
>{\columncolor[HTML]{FFFFC7}}c 
>{\columncolor[HTML]{FFFFC7}}c 
>{\columncolor[HTML]{FFFFC7}}c c
>{\columncolor[HTML]{ECF4FF}}c 
>{\columncolor[HTML]{ECF4FF}}c 
>{\columncolor[HTML]{ECF4FF}}c 
>{\columncolor[HTML]{ECF4FF}}c 
>{\columncolor[HTML]{ECF4FF}}c 
>{\columncolor[HTML]{ECF4FF}}c 
>{\columncolor[HTML]{ECF4FF}}c }
\toprule
\textit{IIIT5k} & \textit{SVT} & \textit{ICDAR2013} & \textit{ICDAR2015} & \textit{SVTP} & \textit{CUTE80} & $\|$ & \textit{Curve} & \textit{Multi-Oriented} & \textit{Artistic} & \textit{Contextless} & \textit{Salient} & \textit{Multi-Words} & \textit{General} 
\end{tabular}}} \\
\toprule
& \multicolumn{2}{c|}{Method}                                              & \multicolumn{6}{c}{\cellcolor[HTML]{FFFFC7}Common Benchmarks (\textit{Com})} & Avg   & \multicolumn{7}{c}{\cellcolor[HTML]{ECF4FF}Union14M-Benchmark (\textit{U14M})} & Avg  & \textit{OST} & \textit{Size}  \\
\midrule
\cellcolor[HTML]{EFEFEF} & CRNN                     & TPAMI~\shortcite{shi2017crnn}            & 95.8     & 91.8     & 94.6     & 84.9     & 83.1     & 91.0    & 90.21 & 48.1    & 13.0    & 51.2    & 62.3    & 41.4    & 60.4   & 68.2   & 49.24  &58.0 & 16.2\\
\cellcolor[HTML]{EFEFEF}& SVTR-B   & IJCAI~\shortcite{duijcai2022svtr}                 & 98.0     & 97.1     & 97.3     & 88.6     & 90.7     & 95.8    & 94.58 & 76.2    & 44.5    & 67.8    & 78.7    & 75.2    & 77.9   & 77.8   & 71.17  &69.6 & 18.1 \\
\multicolumn{1}{c|}{\cellcolor[HTML]{EFEFEF}\multirow{-3}{*}{\begin{tabular}[c]{@{}c@{}}C\\ T\\ C\end{tabular}}} & SVTRv2-B  & ICCV~\shortcite{du2024svtrv2} & \textbf{99.2} & 98.0 & 98.7 & 91.1 & 93.5 & 99.0 & 96.57 & 90.6 & 89.0 & 79.3 & 86.1 & 86.2 & 86.7 & 85.1 & 86.14  & 80.0 & 19.8 
    \\
\hline
\cellcolor[HTML]{EFEFEF}&{DAN}                            & AAAI~\shortcite{wang2020aaai_dan}              & 97.5     & 94.7     & 96.5     & 87.1     & 89.1     & 94.4    & 93.24 & 74.9    & 63.3    & 63.4    & 70.6    & 70.2    & 71.1   & 76.8   & 70.05 & 61.8 &27.7  \\
\cellcolor[HTML]{EFEFEF}&{SEED}                            & CVPR~\shortcite{cvpr2020seed}             & 96.5     & 93.2    & 94.2    & 87.5    & 88.7   & 93.4        & 92.24    & 69.1   & 80.9   & 56.9     & 63.9     & 73.4    & 61.3   & 76.5   & 68.87 & 62.6 &24.0 \\
\cellcolor[HTML]{EFEFEF}&{AutoSTR}                        & ECCV~\shortcite{zhang2020autostr}               & 96.8     & 92.4     & 95.7     & 86.6     & 88.2     & 93.4    & 92.19 & 72.1    & 81.7    & 56.7    & 64.8    & 75.4    & 64.0   & 75.9   & 70.09 & 61.5 &6.0  \\
\cellcolor[HTML]{EFEFEF}&{RoScanner}                      & ECCV~\shortcite{yue2020robustscanner}              & 98.5     & 95.8     & 97.7     & 88.2     & 90.1     & 97.6    & 94.65 & 79.4    & 68.1    & 70.5    & 79.6    & 71.6    & 82.5   & 80.8   & 76.08 & 68.6 &48.0  \\
\cellcolor[HTML]{EFEFEF}&{PARSeq}                        & ECCV~\shortcite{BautistaA22PARSeq}                    & 98.9     & 98.1     & 98.4     & 90.1     & 94.3     & 98.6    & 96.40 & 87.6    & 88.8    & 76.5    & 83.4    & 84.4    & 84.3   & 84.9   & 84.26  & 79.9 &23.8  \\

\cellcolor[HTML]{EFEFEF}&{MAERec}                         & ICCV~\shortcite{jiang2023revisiting}          & \textbf{99.2}               & 97.8                        & 98.2                        & 90.4                        & 94.3                        & 98.3                        & 96.36                        & 89.1                        & 87.1                        & 79.0                        & 84.2                        & 86.3                                 & 85.9                        & 84.6                        & 85.17                   & 76.4     & 35.7                                                \\
\cellcolor[HTML]{EFEFEF}&{LISTER}                        & ICCV~\shortcite{iccv2023lister}               & 98.8     & 97.5     & 98.6     & 90.0     & 94.4     & 96.9    & 96.03 & 78.7    & 68.8    & 73.7    & 81.6    & 74.8    & 82.4   & 83.5   & 77.64 & 77.1 &51.1  \\
\cellcolor[HTML]{EFEFEF}&{CDistNet}                       & IJCV~\shortcite{zheng2024cdistnet}            & 98.7     & 97.1     & 97.8     & 89.6     & 93.5     & 96.9    & 95.59 & 81.7    & 77.1    & 72.6    & 78.2    & 79.9    & 79.7   & 81.1   & 78.62 & 71.8 &43.3 \\
\cellcolor[HTML]{EFEFEF}&{CAM}                            & PR~\shortcite{yang2024class_cam}               & 98.2     & 96.1     & 96.6     & 89.0     & 93.5     & 96.2    & 94.94 & 85.4    & 89.0    & 72.0    & 75.4    & 84.0    & 74.8   & 83.1   & 80.52  & 74.2 & 58.7 \\
\cellcolor[HTML]{EFEFEF}&{OTE}                            & CVPR~\shortcite{Xu_2024_CVPR_OTE}                  & 98.6     & 96.6     & 98.0     & 90.1     & 94.0     & 97.2    & 95.74 & 86.0    & 75.8    & 74.6    & 74.7    & 81.0    & 65.3   & 82.3   & 77.09  & 77.8 &20.3 \\
\cellcolor[HTML]{EFEFEF}&{SMTR}  & AAAI~\shortcite{du2024smtr}    & 99.0  & 97.4 & 98.3 & 90.1 & 92.7          & 97.9 & 95.90          & 89.1 & 87.7 & 76.8 & 83.9 & 84.6 & \textbf{89.3}          & 83.7 & 85.00    & 73.5    & 15.8    \\
\cellcolor[HTML]{EFEFEF}&{IGTR}  & TPAMI~\shortcite{du2024igtr}        &   98.7    &  \textbf{98.4}    &   98.1    &   90.5     & 94.9     &    98.3  & 96.48                                                                       &  90.4     &     91.2      & 77.0      &   82.4          &     84.7    &  84.0                                                      &  84.4       & 84.86  & 76.3 &    24.1    \\
\cellcolor[HTML]{EFEFEF} & \multicolumn{2}{l}{\cellcolor[HTML]{D9E1F4}MDiff4STR-S-AR}     &  \textbf{99.2} &	98.1 &	98.6 &	90.4 &	94.9 &	98.3 &	96.58 &	91.8 &	94.2 &	79.8 &	85.4 &	\textbf{88.1} &	86.9 &	85.9 &	87.42  & 79.3 &   18.9   \\
\multicolumn{1}{c|}{\cellcolor[HTML]{EFEFEF}\multirow{-14}{*}{\begin{tabular}[c]{@{}c@{}}A\\ R \\M\end{tabular}}} & \multicolumn{2}{l}{\cellcolor[HTML]{D9E1F4}MDiff4STR-B-AR}       & 99.1 &	98.1 &	\textbf{99.2} &	\textbf{91.6} &	96.0 &	98.6 &	97.09 &	93.6 	&\textbf{94.6} &	81.9 &	\textbf{86.5} &	\textbf{88.1} 	&89.0 &	86.7 &	\textbf{88.63}  & 82.5 &   31.9   \\

\hline
 \cellcolor[HTML]{EFEFEF}&{VisionLAN}                      & ICCV~\shortcite{Wang_2021_visionlan}           & 98.2     & 95.8     & 97.1     & 88.6     & 91.2     & 96.2    & 94.50 & 79.6    & 71.4    & 67.9    & 73.7    & 76.1    & 73.9   & 79.1   & 74.53  & 66.4 &32.9  \\
\cellcolor[HTML]{EFEFEF}&{MGP-STR}                        & ECCV~\shortcite{mgpstr}                    & 97.9 &	97.8 &	97.1 	&89.6 	&95.2 	&96.9 	&95.75 	&85.2 	&83.7 	&72.6 	&75.1 	&79.8 	&71.1 	&83.1 	&78.65   & 78.7 & 148 \\
\cellcolor[HTML]{EFEFEF}&{LPV}                          & IJCAI~\shortcite{ijcai2023LPV}                 & 98.6     & 97.8     & 98.1     & 89.8     & 93.6     & 97.6    & 95.93 & 86.2    & 78.7    & 75.8    & 80.2    & 82.9    & 81.6   & 82.9   & 81.20  & 77.7 & 30.5 \\
\cellcolor[HTML]{EFEFEF}&{CPPD}                         & TPAMI~\shortcite{du2023cppd}                  & 99.0     & 97.8     & 98.2     & 90.4     & 94.0     & \textbf{99.0}    & 96.40 & 86.2    & 78.7    & 76.5    & 82.9    & 83.5    & 81.9   & 83.5   & 81.91 & 79.6 &27.0  \\

\cellcolor[HTML]{EFEFEF} &\multicolumn{2}{l}{\cellcolor[HTML]{D9E1F4}MDiff4STR-S-PD}          & 98.6 &	98.1 &	98.5 &	89.1 &	93.8 &	97.9 &	96.00 &	90.4 &	92.3 &	77.9 &	83.2 &	86.0 &	80.5 &	84.6 &	84.98   & 77.4 &    18.9  \\
\multicolumn{1}{c|}{\cellcolor[HTML]{EFEFEF}\multirow{-6}{*}{\begin{tabular}[c]{@{}c@{}}P\\ D \\M\end{tabular}}}  &\multicolumn{2}{l}{\cellcolor[HTML]{D9E1F4}MDiff4STR-B-PD}          & 98.9 	&98.3 &	98.9 	& 90.8 	&95.7 &	98.6 	&96.88 	&92.6 &	93.6 &	79.0 	&84.7 &	86.7 &	84.5 &	85.7 &	86.69   & 81.3 &    31.9  \\
\hline
\cellcolor[HTML]{EFEFEF} &{SRN}                            & CVPR~\shortcite{yu2020srn}              & 97.2    & 96.3      & 97.5     & 87.9     & 90.9     & 96.9    & 94.45 & 78.1    & 63.2    & 66.3    & 65.3    & 71.4    & 58.3   & 76.5   & 68.43 & 64.6 &51.7  \\
\cellcolor[HTML]{EFEFEF}&{ABINet}                         & CVPR~\shortcite{fang2021abinet}            & 98.5     & 98.1     & 97.7     & 90.1     & 94.1     & 96.5    & 95.83 & 80.4    & 69.0    & 71.7    & 74.7    & 77.6    & 76.8   & 79.8   & 75.72 & 75.0 &36.9  \\
\cellcolor[HTML]{EFEFEF}&{MATRN}                          & ECCV~\shortcite{MATRN}            & 98.8     & 98.3     & 97.9     & 90.3     & 95.2     & 97.2    & 96.29 & 82.2    & 73.0    & 73.4    & 76.9    & 79.4    & 77.4   & 81.0   & 77.62  & 77.8 &44.3 \\
\cellcolor[HTML]{EFEFEF}&{BUSNet}                         & AAAI~\shortcite{Wei_2024_busnet}                   & 98.3     & 98.1     & 97.8     & 90.2     & 95.3     & 96.5    & 96.06 & 83.0    & 82.3    & 70.8    & 77.9    & 78.8    & 71.2   & 82.6   & 78.10  & 78.7 &32.1 \\

\cellcolor[HTML]{EFEFEF} &\multicolumn{2}{l}{\cellcolor[HTML]{D9E1F4}MDiff4STR-S-Re}              &  99.0 &	98.1 &	98.0 &	90.0 &	94.3 &	97.9 &	96.22 &	92.0 &	93.9 &	80.4 &	84.1 &	87.1 &	85.2 &	85.8 &	86.93   & 81.7 &    18.9    \\
\multicolumn{1}{c|}{\cellcolor[HTML]{EFEFEF}\multirow{-6}{*}{\begin{tabular}[c]{@{}c@{}}R\\ e \\ M\end{tabular}}}&\multicolumn{2}{l}{\cellcolor[HTML]{D9E1F4}MDiff4STR-B-Re}              &   \textbf{99.2} &	98.3 	&98.6 &	91.3 &	96.7& 	98.6 &	97.12 &	93.4 &	94.3 &	\textbf{82.2} &	86.0 &	87.7 &	86.8 &	86.7 	& 88.16  & 83.7 &    31.9   \\

\hline
\cellcolor[HTML]{EFEFEF}&\multicolumn{2}{l}{\cellcolor[HTML]{D9E1F4}MDiff4STR-S-LC}              &  99.0 &	98.3 &	98.4 &	90.2 &	94.9 &	97.9 &	96.44 &	91.8 &	94.1 & 	80.1 &	85.1 &	87.3 &	84.3 &	85.9 &	86.95 &	81.3  &    18.9   \\
\cellcolor[HTML]{EFEFEF}&\multicolumn{2}{l}{\cellcolor[HTML]{D9E1F4}MDiff4STR-B-LC}              &   \textbf{99.2} &	98.3 &	99.1 &	\textbf{91.6} &	\textbf{97.1} &	98.6 &	97.29 &	\textbf{93.7} &	94.4 &	82.0 &	86.0 &	87.9 &	87.7 &	\textbf{86.8} &	88.37 & 	84.2  &    31.9  \\

\cellcolor[HTML]{EFEFEF}&\multicolumn{2}{l}{\cellcolor[HTML]{D9E1F4}MDiff4STR-S-BLC}              &   99.0 & 	98.3 & 	98.4 & 	90.2 & 	94.9 & 	97.6 & 	96.38 & 	91.8 & 	94.0 & 	80.2 & 	85.1 & 	87.3 & 	84.8 & 	85.9 & 	87.03 & 	81.4 &    18.9  \\

\cellcolor[HTML]{EFEFEF}&\multicolumn{2}{l}{\cellcolor[HTML]{D9E1F4}MDiff4STR-B-BLC}              &   \textbf{99.2} &	98.3 &	99.1 &	\textbf{91.6} &	\textbf{97.1} &	98.6 &	\textbf{97.30} &	\textbf{93.7} &	94.4 &	82.1 &	86.1 &	87.7 &	88.3 &	\textbf{86.8} &	88.44 & 	\textbf{84.3} &    31.9   \\

\bottomrule
\end{tabular}}
\caption{All the models are trained on \textit{U14M-Filter} from scratch. \textit{Size} denotes the number of parameters of the model ($\times 10^6$).}
\label{tab:sota}
\end{table*}

\noindent\textbf{Effectiveness of the MDM in STR}. The vanilla MDM (second part of Tab.~\ref{tab:mask_reflect}) is trained using only a random mask strategy. During inference, we explore five different denoising inference paradigms (MDiff-PD/AR/Re/LC/BLC, detailed in Sec.~2.2). Among these, The results show that MDiff-LC and MDiff-BLC, which represent entirely new decoding strategies distinct from previous STR methods, both achieve competitive results comparing with the existing STR methods in terms of accuracy (see Tab.~\ref{tab:sota}), despite trailing behind the ARM$_{base}$. Nonetheless, MDiff-BLC delivers a substantial advantage in efficiency, achieving an average inference speed 3× faster than ARM$_{base}$. These findings demonstrate the potential of the MDM for the STR task.

\noindent\textbf{Effectiveness of Training Mask Strategies}. As shown in the third part of Tab.~\ref{tab:mask_reflect}, incorporating six training mask strategies in Fig.~\ref{fig:mask7} with the random mask strategy, all five denoising paradigms of MDiff4STR exhibit significant performance gains. On the \textit{Com}, \textit{U14M}, and \textit{OST} the average improvements are 0.43\%, 1.68\%, and 2.17\%, respectively. The \textit{U14M} and \textit{OST}, the more challenging test sets, benefit most from the six training mask strategies, which effectively bridge the noising gap between training and inference. Furthermore, as shown in the bottom part of Tab.~\ref{tab:mask_reflect}, all masks contribute positively when added gradually. Among them, the full mask strategy is the most effective, which can be explained by its role as the initial denoising step, laying a critical foundation for all subsequent denoising stages.

\noindent\textbf{Effectiveness of Token-Replacement Noise.} As shown in the fourth part of Tab.~\ref{tab:mask_reflect}, the token-replacement noise mechanism (TRN) further improves performance, boosting average accuracy over vanilla MDM by 0.91\%, 2.93\%, and 4.58\% on \textit{Com}, \textit{U14M}, and \textit{OST}, respectively. This allows MDiff4STR-LC and MDiff4STR-BLC to outperform the ARM$_{base}$ using only three denoising steps. In addition, as shown in Fig.~\ref{fig:case}, when the model encounters high-confidence score but incorrect predictions (e.g., erroneous tokens that were not re-masked), MDiff4STR can recognize and correct these errors in subsequent denoising steps. This excellent error correction ability and denoising stability is precisely benefiting from our proposed TRN.

\noindent\textbf{Effectiveness of the MDM's Omnidirectional Language Modeling.} As a novel paradigm, MDiff4STR shows its most representative strength on \textit{OST}, where its accuracy improves from 81.03\% (ARM$_{base}$) to 84.25\%, a 3.22\% gain, and a significant 5.27\% increase over the ReM$_{base}$ (78.98\%). The \textit{OST}, which focuses on occluded scenes, is designed to evaluate a model’s contextual modeling ability to infer missing or obstructed characters. Furthermore, the last three samples in Fig.~\ref{fig:case} also illustrate that MDiff4STR outperforms ARM and ReM when encountering occluded and artistic text that requires contextual reasoning for correct recognition. These results indicate that MDM’s omnidirectional contextual modeling ~\cite{ShiHWDT24_simmaskdiff_dis,SahooASGMCRK24_simmaskdiff} offers a clear advantage over traditional unidirectional auto-regressive methods (e.g., ARM) and bidirectional refinement models (e.g., ReM).

\begin{table}[t]\footnotesize
\centering
\setlength{\tabcolsep}{1.9pt}{
\begin{tabular}{r|cccccc|c|c}
\toprule
 Method      &  \multicolumn{6}{c|}{Common Benchmarks (\textit{Com})}    & Avg & \textit{OST}  \\
\hline
E$^2$STR~\shortcite{Zhao_2024_CVPR_E2STR}   & 99.2 & 98.6 & 98.7 & 93.8 & 96.7 & 99.3 & 97.71 & 80.7  \\
 VL-Reader~\shortcite{zhong_2024_acmmm_vlreader}  & \textbf{99.6} & 99.1 & 98.7 & 92.6 & 97.5 & 99.3 & 97.80 & 86.2 \\
 CLIP4STR~\shortcite{zhao_2025_tip_clip4str}    & 99.4 & 98.6 & 98.3 & 90.8 & \textbf{97.8} & 99.0   & 97.32 & 82.8  \\
 DPTR~\shortcite{zhao_2024_acmmm_dptr}    & 99.5 & \textbf{99.2} & 98.5 & 91.8 & 97.1 & 98.6   & 97.45 & -  \\
 IGTR~\shortcite{du2024igtr}   & 99.2      &   98.3   &   98.8   &  92.0    &  96.8    &   99.0   & 97.34 & 86.5   \\
 SVTRv2-B~\shortcite{du2024svtrv2}             & 99.2 &	98.6 &	98.8 &	93.8 &	97.2 &	99.4 &  97.83 &	86.9 \\
 \hline
\cellcolor[HTML]{D9E1F4}MDiff4STR-BLC   & 99.5 &	98.5 &	\textbf{98.9} &	\textbf{94.1} &	97.4 &	\textbf{99.7} &  \textbf{98.02} &	\textbf{87.4} \\
\bottomrule
\end{tabular}}
\caption{Quantitative comparison of MDiff4STR-B with the advanced methods experienced large-scale pretraining.}
\label{tab:pretraining}
\end{table}

\begin{table}[t]\footnotesize
\centering
\setlength{\tabcolsep}{4.0pt}{
\begin{tabular}{r|cccc|c|c}
\toprule
Method        & \textit{Scene}         & \textit{Web}           & \textit{Doc} & \textit{HW} & Avg   & \textit{Size}         \\
\hline

CRNN~\shortcite{shi2017crnn} & 63.8 & 68.2 & 97.0 & 
 46.1 & 68.76 & 19.5  \\ 
 SVTR-B~\shortcite{duijcai2022svtr} & 77.9 	& 78.7 &	99.2 	&62.1 	&79.49  & 19.8 \\
 DCTC~\shortcite{Zhang_Lu_Liao_Huang_Li_Wang_Peng_2024_DCTC}     & 73.9          & 68.5          & 99.4          & 51.0          & 73.20      & 40.8     \\
SVTRv2-B~\shortcite{du2024svtrv2}  &  83.5 &	83.3 &	99.5 &	67.0 &	83.31    &22.5            \\
\hline
ASTER~\shortcite{shi2019aster}        & 61.3          & 51.7          & 96.2          & 37.0            & 61.55      & 27.2   \\
MORAN~\shortcite{pr2019MORAN}        & 54.6          & 31.5          & 86.1          & 16.2          & 47.10        & 28.5   \\
SAR~\shortcite{li2019sar}          & 59.7          & 58.0            & 95.7          & 36.5          & 62.48      & 27.8   \\
SEED~\shortcite{cvpr2020seed}         & 44.7          & 28.1          & 91.4          & 21.0            & 46.30        & 36.1   \\
MASTER~\shortcite{pr2021MASTER}       & 62.8          & 52.1          & 84.4          & 26.9          & 56.55     & 62.8   \\
TransOCR~\shortcite{cvpr2021TransOCR}     & 71.3          & 64.8          & 97.1          & 53.0            & 71.55      & 83.9   \\
PARSeq~\shortcite{BautistaA22PARSeq} & 84.2 & 82.8 & 99.5 & 63.0 & 82.37 &28.9 
\\
CCR-CLIP~\shortcite{yuICCV2023clipctr}     & 71.3          & 69.2          & 98.3          & 60.3          & 74.78        & 62.0     \\
CAM~\shortcite{yang2024class_cam}     & 76.0          & 69.3          & 98.1          & 59.2          & 76.80        & 135     \\
MAERec~\shortcite{jiang2023revisiting} & 84.4 & 83.0 & 99.5 & 65.6 & 83.13  & 40.8\\
LISTER~\shortcite{iccv2023lister} & 79.4 & 79.5 & 99.2 & 58.0 & 79.02 & 55.0 \\
DPTR~\shortcite{zhao_2024_acmmm_dptr} & 80.0 & 79.6 & 98.9 & 64.4  & 80.73 & 68.0 \\
IGTR-AR~\shortcite{du2024igtr}     & 82.0 & 81.7 & 99.5 &  63.8 &  81.74   & 29.2 \\
SMTR~\shortcite{du2024smtr}       & 83.4          &   83.0            &  99.3             &    65.1           &  82.68     &   20.8      \\
\cellcolor[HTML]{D9E1F4}MDiff4STR-S-AR  & 84.4          &   83.8            &  99.5             &    67.1           &  83.71 &    23.9          \\
\cellcolor[HTML]{D9E1F4}MDiff4STR-B-AR  & 85.1          &   84.2            &  \textbf{99.6}             &    \textbf{67.3}           &  84.04 &    36.9              \\
\hline
CPPD~\shortcite{du2023cppd} & 82.7 	& 82.4 	& 99.4 	& 62.3 	&81.72 & 32.1 \\
\cellcolor[HTML]{D9E1F4}MDiff4STR-S-PD  & 82.4          &   82.5            &  99.4             &    60.9           &  81.31 &    23.9               \\
\cellcolor[HTML]{D9E1F4}MDiff4STR-B-PD  & 83.4          &   82.2            &  99.5             &    61.8           &  81.96 &    36.9              \\
\hline
ABINet~\shortcite{fang2021abinet}       & 66.6          & 63.2          & 98.2          & 53.1          & 70.28        & 53.1   \\

\cellcolor[HTML]{D9E1F4}MDiff4STR-B-Re  & 84.6          &   83.9            &  99.5             &    65.4           &  83.39 &    23.9              \\

\cellcolor[HTML]{D9E1F4}MDiff4STR-B-Re  & 85.6          &   84.7            &  \textbf{99.6}             &    67.0           &  84.23 &    36.9              \\
\hline
\cellcolor[HTML]{D9E1F4}MDiff4STR-S-LC & 85.2 & 84.1 & \textbf{99.6} & 66.0 & 83.72 &    23.9   \\
\cellcolor[HTML]{D9E1F4}MDiff4STR-B-LC & 85.6 & \textbf{84.8} & \textbf{99.6} & 66.5 & 84.11 &    36.9  \\
\cellcolor[HTML]{D9E1F4}MDiff4STR-S-BLC & 85.2 & 84.1 & \textbf{99.6} & 66.7 & 83.89 &    23.9  \\
\cellcolor[HTML]{D9E1F4}MDiff4STR-B-BLC & \textbf{85.7} & 84.7 & \textbf{99.6} & 67.0 & \textbf{84.25} &    36.9  \\
\bottomrule
\end{tabular}}
\caption{Results on Chinese text dataset.}
\label{tab:ch_all}
\end{table}

\subsection{Comparison with State-of-the-arts}

To facilitate a systematic comparison, we categorize existing STR methods into four decoding paradigms: Connectionist temporal classification (CTC)-based model \cite{CTC}, ARM, PDM, and ReM, as summarized in Tab.~\ref{tab:sota}. To eliminate the influence of model size, we report results for two MDiff4STR variants: MDiff4STR-S (18.9M parameters) and MDiff4STR-B (31.9M parameters). Under the ARM, PDM, and ReM paradigms, MDiff4STR-S and MDiff4STR-B consistently outperform competing models of similar size on \textit{Com}, \textit{U14M}, and \textit{OST}. Notably, the one-step denoising version, MDiff4STR-B-PD achieves state-of-the-art results, outperforming previous bests by 0.31\%, 0.55\%, and 1.30\% on the three benchmarks, respectively. Further improvements are observed with the dedicated decoding strategy MDiff4STR-B-BLC, which outperforms the previous best results by 0.73\%, 2.30\%, and 4.30\% on \textit{Com}, \textit{U14M}, and \textit{OST}, respectively. In particular, the 4.30\% gain on \textit{OST} demonstrates the effectiveness of MDiff4STR’s omnidirectional language modeling in capturing complex contextual dependencies, offering a novel and powerful decoding paradigm for STR.

To explore the potential of MDiff4STR in leveraging large-scale pretraining, we first pretrain it on synthetic datasets~\cite{Synthetic,jaderberg14synthetic} and then fine-tune it on the real-world dataset \textit{U14M-Filter}. As shown in Tab.~\ref{tab:pretraining}, MDiff4STR clearly benefits from pretraining: compared to training from scratch (in Tab.~\ref{tab:sota}), accuracy improves by 0.72\% on \textit{Com} and a significant 3.10\% on \textit{OST}. Compared with other state-of-the-art pretrained models, MDiff4STR achieves the highest average accuracy on both \textit{Com} (98.02\%) and \textit{OST} (87.4\%), further validating the strength of its omnidirectional language modeling framework for robust and generalizable STR.

In Tab.~\ref{tab:ch_all}, we present the results of MDiff4STR on BCTR \cite{chen2021benchmarking}, a challenging Chinese text recognition benchmark. Compared to English, Chinese text poses greater complexity due to intricate stroke structures and a significantly larger character set. Despite these challenges, MDiff4STR outperforms the previous state-of-the-art by 1.3\%, 1.5\%, and 0.1\% on the \textit{Scene}, \textit{Web}, and \textit{Document} (\textit{Doc}) subsets, respectively, and achieves comparable performance on the \textit{Hand-Writing} (\textit{HW}) subset. These results highlight the robustness and adaptability of MDiff4STR to non-Latin scripts, further validating its generalizability in multilingual text recognition tasks.

\section{Conclusion}

In this paper, we introduced the MDM into the STR task and proposed a novel STR framework, MDiff4STR. Leveraging MDM’s omnidirectional language modeling capability, MDiff4STR surpasses the previous prominent ARMs and BERT-like models. To overcome two key challenges in applying the MDM to STR, the noising gap between training and inference and the overconfidence in predictions during inference, we proposed the training noise strategy that aligns with inference behavior and the token-replacement noise mechanism. Extensive experiments demonstrate that MDiff4STR can flexibly support multiple decoding paradigms and achieves state-of-the-art results across a wide range of challenging scenarios, including regular, irregular, occluded, and Chinese text, with or without the use of pretraining. Furthermore, its dedicated low-confidence denoising inference surpasses ARMs with only three denoising steps, establishing a novel and effective paradigm for STR. These findings highlight the potential of MDiff4STR to advance the STR field and offer valuable insights for future research.

\section*{Acknowledgments}

This work was supported by National Natural Science Foundation of China (Nos. 62427819, 62576026)

\bibliography{aaai2026}
\appendix
\clearpage



\begin{table*}[t]\footnotesize
\centering
\setlength{\tabcolsep}{4.5pt}{
\begin{tabular}{c|c|c|c}
\toprule
Language Model & Conditional Probability  & Context Type & Supported Decoding Paradigm  \\
\midrule
ARM & $P(x_t \mid x_1, x_2, \dots, x_{t-1},  \mathbf{F}_v)$       & Left to one unidirectional   & Auto-Regressive Decoding \\
\midrule
PDM & $P(x_t \mid \mathbf{F}_v)$      & None  & Parallel Decoding \\
\midrule
ReM & $P(x_t \mid x_1, \dots, x_{t-1}, x_{t+1}, \dots, x_{T}, \mathbf{F}_v)$     & Left and right to one bidirectional  & Refinement Decoding  \\
\midrule
PARM & \begin{tabular}[c]{@{}c@{}}\(\pi = \text{RandomShuffle}(\{1, 2, \dots, T\})\), \\ \(P(x_{\pi_t} \mid x_{\pi_1}, x_{\pi_2}, \dots, x_{\pi_{t-1}}, \mathbf{F}_v)\)\end{tabular}   & Any to one unidirectional & \begin{tabular}[c]{@{}c@{}}Auto-Regressive Decoding \\ Parallel Decoding \\Refinement Decoding\end{tabular}  \\
\midrule
MDM & \begin{tabular}[c]{@{}c@{}}\(\pi = \text{RandomShuffle}(\{1, 2, \dots, T\})\), \\ \(P(x_{\pi_t},\dots,x_{\pi_T} \mid x_{\pi_1}, x_{\pi_2}, \dots, x_{\pi_{t-1}}, \mathbf{F}_v)\)\end{tabular} & Any to any omnidirectional &  \begin{tabular}[c]{@{}c@{}}Auto-Regressive Decoding \\ Parallel Decoding \\Refinement Decoding \\Low-Confidence Remask\\ Block Low-Confidence Remask \end{tabular} \\
\bottomrule
\end{tabular}}
\caption{Comparison of language models in terms of conditional probability, context type, and decoding strategy. Here, $x = \{x_1, x_2, \dots, x_T\}$ denotes the target token sequence of length $T$, where each $x_t$ represents the token at position $t$, 
$\pi$ denotes a random permutation of token positions, and $\mathbf{F}_v$ represents visual features. }
\label{tab:lm}
\end{table*}

\section{Related Work}

Scene Text Recognition (STR), as a typical vision-language task, often heavily relies on linguistic context when dealing with complex texts that are difficult to recognize directly, such as those with occlusion, blur, or artistic fonts.

\subsubsection{Auto-Regressive Model (ARM) for STR:}
Most efforts~\cite{shi2016rare,shi2019aster,Sheng2019nrtr,li2019sar,yue2020robustscanner,jiang2023revisiting,xie2022toward_cornertrans,zheng2024cdistnet,Xu_2024_CVPR_OTE,yang2024class_cam,zhou2024cff,du2024igtr,du2024smtr} to incorporate language modeling capabilities into STR were inspired by the sequence-to-sequence frameworks commonly used in machine translation and speech recognition \cite{seqtoseq,NIPS2017_attn,lstm}. These methods are typically based on ARMs, which naturally introduce left-to-right \textit{unidirectional language modeling} by iteratively predicting characters one by one. However, the significant advancements brought by ARMs in STR, their inherently low inference efficiency has long been a major limitation. 

\subsubsection{Parallel Decoding Model (PDM) and Refinement Model (ReM) for STR:}
Recent years have witnessed the emergence of Non-Auto-Regressive (NAR) methods ~\cite{wang2020aaai_dan,yu2020srn,fang2021abinet,Wang_2021_visionlan,mgpstr,du2023cppd,ijcai2023LPV,MATRN,Wei_2024_busnet}. Among them, Parallel Decoding Models~\cite{wang2020aaai_dan,qiao2021pimnet,Wang_2021_visionlan,mgpstr,du2023cppd,ijcai2023LPV} have gained attention for their ability to predict all characters simultaneously in a single forward pass, thereby greatly improving inference speed. However, this ``brute-force" decoding paradigm tends to neglect the linguistic dependencies between characters, often resulting in a notable decline in recognition accuracy. To mitigate this shortcoming, a growing body of research~\cite{yu2020srn,fang2021abinet,BautistaA22PARSeq,MATRN,Wang_2021_visionlan,du2023cppd,Wei_2024_busnet} has focused on incorporating language context into NAR frameworks to enhance recognition performance. One of the most representative methods ~\cite{yu2020srn,fang2021abinet,MATRN,Wei_2024_busnet} in this direction is the refinement model (ReM) based on BERT-like architectures. These methods build upon PDMs by introducing a cloze-style language model that iteratively refines the initial predictions using left and right to one \textit{bidirectional linguistic modeling}, striking a balance between recognition accuracy and computational efficiency. 
Although these methods alleviate, the weak language modeling capability inherent in NAR architectures, they fundamentally reflect different strategies for balancing recognition accuracy and inference speed. Experimental results suggest that the recognition accuracy of some of these NAR-based models still lags behind that of state-of-the-art ARMs.

In particular, PARSeq~\cite{BautistaA22PARSeq} and IGTR~\cite{du2024igtr} represent distinct advancements that unify multiple decoding paradigms, i.e., ARM, PDM, and ReM, within a single model. PARSeq leverages permutation auto-regressive language modeling to enable \textit{language modeling from the any direction to one token}, thereby offering the potential for flexible switching among the three decoding paradigms. Specifically, permutation auto-regressive model (PARM) is an autoregressive language modeling paradigm in which the generation order of tokens is determined by a sampled permutation of the sequence indices, rather than a fixed left-to-right or right-to-left order. This approach allows the model to capture richer contextual dependencies while maintaining an autoregressive training objective. IGTR, on the other hand, formulates STR as an instruction learning task. By flexibly adjusting the input instructions, it achieves unified modeling of different decoding paradigms, demonstrating high adaptability and extensibility.

\begin{table*}[t]\footnotesize
\centering
\begin{tabular}{c|c|c|c|c|c|c}
\toprule
Models & $\left[D_0, D_1, D_2 \right]$ & $[N_1, N_2, N_3]$ & Heads   & Permutation & $N$ & \textit{Size} \\
\midrule
MDiff4STR-S & {[}128,256,384{]}  & {[}3,6,3{]}      & {[}4,8,12{]}  & $[L]_6  {[G]_6 }$ & $3$ & 18.9 \\
MDiff4STR-B & {[}128,256,384{]} & {[}6,6,6{]}      & {[}4,8,12{]} & $[L]_8 { [G]_{10}}$ & $6$ & 31.9 \\
\bottomrule
\end{tabular}
\caption{Architecture specifications of MDiff4STR variants.}
\label{tab:booktab1}
\end{table*}

In Tab.~\ref{tab:lm}, we provide a systematic comparison of the previous language modeling approaches across three key dimensions: (1) the formulation of conditional probabilities, (2) the type of contextual information the model utilizes during generation; and (3) the supported decoding paradigms. In contrast to the previous methods, the MDM introduces a novel perspective. It proposes an \textit{omnidirectional language modeling} capability that subsumes the unidirectional left-to-one  modeling in ARMs, the bidirectional left and right-to-one modeling in ReMs, and the any-to-one directional modeling used in PARSeq. Building on this advantage, MDiff4STR not only supports the three mainstream decoding paradigms (AR, PD, and Re), but also introduces two confidence-based adaptive decoding mechanism (LC and BLC). Both mechanisms allow MDiff4STR to fully exploit its omnidirectional language modeling capacity during iterative denoising, surpassing the recognition accuracy of ARMs within three iterations. Consequently, MDiff4STR establishes a novel STR paradigm that combines high accuracy with high efficiency, demonstrating both strong practical potential and significant theoretical value.

\begin{figure*}[t]
  \centering
\includegraphics[width=0.98\textwidth]{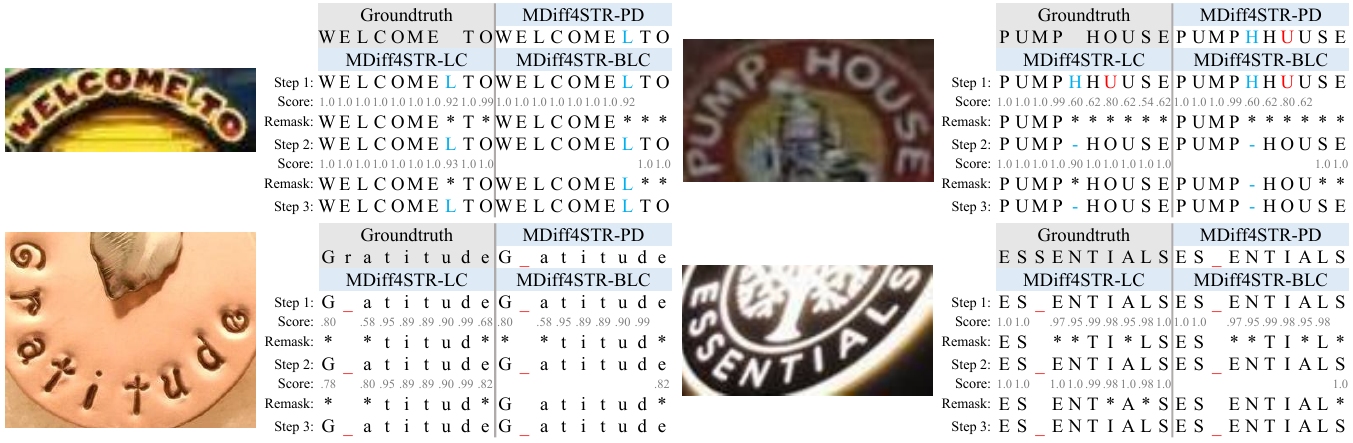} 
  \caption{Bad samples with redundant and missing recognition results of MDiff4STR. Red characters indicate recognition errors. Blue characters and red underlines indicate redundant and missing characters, respectively.}
  \label{fig:badcase}
\end{figure*}

\subsubsection{Mask Diffusion Model (MDM):} Diffusion models~\cite{DenoisingM,latentDiff} initially garnered significant attention in the field of image generation, and recently have emerged as promising alternatives to ARMs in the domain of text generation, sparking considerable research interest. Sahoo et al. ~\cite{SahooASGMCRK24_simmaskdiff} introduced the MDM, which reformulates text generation as a masked token prediction task. This formulation enables the diffusion process to be effectively adapted to general-purpose language tasks. Subsequent studies ~\cite{ShiHWDT24_simmaskdiff_dis,SahooASGMCRK24_simmaskdiff} have demonstrated that MDM, as a form of Omnidirectional Language Modeling, possesses the potential to surpass traditional ARMs in both modeling flexibility and performance. Building on these insights, LLaDA~\cite{llada} successfully applied the MDM framework to large-scale language models, achieving notable performance improvements and drawing widespread attention. Extending this line of work further, LLaDA-V~\cite{lladav} integrated MDM into multimodal large language models, significantly advancing the model's cross-modal capabilities. 

Inspired by these progress in the MDM, we pioneer the integration of the MDM framework into the STR task. To this end, we develop MDiff4STR, a tailored version of MDM for STR. Experiments show that MDiff4STR achieves superior recognition accuracy and efficiency compared to ARMs, even with three denoising steps. These results underscore the potential of MDM as a novel and efficient paradigm for STR.

\section{MDiff4STR Variants}

Following the SVTRv2~\cite{du2024svtrv2}, we change the hyper-parameters of the model structure to form two variants, i.e., MDiff4STR-S (Small), MDiff4STR-B (Base). These hyper-parameters include the depth of channel ($D_i$) and the number of heads at each stage, the number of mixing blocks ($N_i$), their permutation and the number of layers of the mask diffusion decoder $N$. Their detail configurations are shown in Tab.~\ref{tab:booktab1}. The depth of channel ($D_i$) and the number of heads at each stage, the number of mixing blocks ($N_i$), and their permutation are used to control the structure of the visual encoder SVTRv2~\cite{du2024svtrv2}. SVTRv2 is a three-stage, each stage using a number of local mixing or global mixing blocks hierarchical backbone network. In Tab.~\ref{tab:booktab1}, $[L]_m { [G]_n}$ denotes that the first \textit{m} mixing blocks in SVTRv2 utilize local mixing, while the last \textit{n} mixing blocks employ global mixing.

\begin{table*}[t]\footnotesize
\centering
\setlength{\tabcolsep}{3.5pt}{
\begin{tabular}{c|ll|ccccccc|cccccccc|c}
\multicolumn{19}{c}{\setlength{\tabcolsep}{2pt}{\begin{tabular}{
>{\columncolor[HTML]{FFFFC7}}c 
>{\columncolor[HTML]{FFFFC7}}c 
>{\columncolor[HTML]{FFFFC7}}c 
>{\columncolor[HTML]{FFFFC7}}c 
>{\columncolor[HTML]{FFFFC7}}c 
>{\columncolor[HTML]{FFFFC7}}c c
>{\columncolor[HTML]{ECF4FF}}c 
>{\columncolor[HTML]{ECF4FF}}c 
>{\columncolor[HTML]{ECF4FF}}c 
>{\columncolor[HTML]{ECF4FF}}c 
>{\columncolor[HTML]{ECF4FF}}c 
>{\columncolor[HTML]{ECF4FF}}c 
>{\columncolor[HTML]{ECF4FF}}c }
\toprule
\textit{IIIT5k} & \textit{SVT} & \textit{ICDAR2013} & \textit{ICDAR2015} & \textit{SVTP} & \textit{CUTE80} & $\|$ & \textit{Curve} & \textit{Multi-Oriented} & \textit{Artistic} & \textit{Contextless} & \textit{Salient} & \textit{Multi-Words} & \textit{General} 
\end{tabular}}} \\
\toprule
& \multicolumn{2}{c|}{Method}       & \multicolumn{6}{c}{\cellcolor[HTML]{FFFFC7}Common Benchmarks (\textit{Com})} & Avg   & \multicolumn{7}{c}{\cellcolor[HTML]{ECF4FF}Union14M-Benchmark (\textit{U14M})} & Avg  & \textit{Size} \\
\midrule

\cellcolor[HTML]{EFEFEF}&CRNN        & TPAMI~\shortcite{shi2017crnn} & 82.9 & 81.6 & 91.1 & 69.4 & 70.0 & 65.5 & 76.75 & 7.50  & 0.90  & 20.7 & 25.6 & 13.9 & 25.6 & 32.0 & 18.03 & 8.30 \\
\cellcolor[HTML]{EFEFEF}&SVTR     & IJCAI~\shortcite{duijcai2022svtr}    & 96.0 & 91.5 & 97.1 & 85.2 & 89.9 & 91.7 & 91.90 & 69.8 & \textbf{37.7} & 47.9 & 61.4 & 66.8 & 44.8 & 61.0 & 55.63 & 24.6 \\
\cellcolor[HTML]{EFEFEF}&DCTC        & AAAI~\shortcite{Zhang_Lu_Liao_Huang_Li_Wang_Peng_2024_DCTC}        & 96.9 & 93.7 & 97.4 & 87.3 & 88.5 & 92.3 & 92.68 & -    & -    & -    & -    & -    & -    & -    & -     & 40.8 \\
\multicolumn{1}{c|}{\cellcolor[HTML]{EFEFEF}\multirow{-4}{*}{\begin{tabular}[c]{@{}c@{}}C\\ T\\ C\end{tabular}}}&SVTRv2       &   ICCV~\shortcite{du2024svtrv2}          &97.7 &94.0 &97.3 &88.1 &91.2 &95.8 &94.02 &74.6 &25.2 &57.6 &\textbf{69.7} &77.9 &68.0 &66.9 &62.83 &19.8 \\
\midrule
\cellcolor[HTML]{EFEFEF}&ASTER        & TPAMI~\shortcite{shi2019aster}   & 93.3 & 90.0 & 90.8 & 74.7 & 80.2 & 80.9 & 84.98 & 34.0 & 10.2 & 27.7 & 33.0 & 48.2 & 27.6 & 39.8 & 31.50 & 27.2 \\
\cellcolor[HTML]{EFEFEF}&NRTR         & ICDAR~\shortcite{Sheng2019nrtr}      & 90.1 & 91.5 & 95.8 & 79.4 & 86.6 & 80.9 & 87.38 & 31.7 & 4.40  & 36.6 & 37.3 & 30.6 & 54.9 & 48.0 & 34.79 & 31.7 \\
\cellcolor[HTML]{EFEFEF}&MORAN       & PR~\shortcite{pr2019MORAN}       & 91.0 & 83.9 & 91.3 & 68.4 & 73.3 & 75.7 & 80.60 & 8.90  & 0.70  & 29.4 & 20.7 & 17.9 & 23.8 & 35.2 & 19.51 & 17.4     \\
\cellcolor[HTML]{EFEFEF}&SAR         & AAAI~\shortcite{li2019sar}  & 91.5 & 84.5 & 91.0 & 69.2 & 76.4 & 83.5 & 82.68 & 44.3 & 7.70  & 42.6 & 44.2 & 44.0 & 51.2 & 50.5 & 40.64 & 57.7 \\
\cellcolor[HTML]{EFEFEF}&DAN        & AAAI~\shortcite{wang2020aaai_dan}  & 93.4 & 87.5 & 92.1 & 71.6 & 78.0 & 81.3 & 83.98 & 26.7 & 1.50  & 35.0 & 40.3 & 36.5 & 42.2 & 42.1 & 32.04 & 27.7     \\

\cellcolor[HTML]{EFEFEF}&SEED        & CVPR~\shortcite{cvpr2020seed}     &  93.8    &  89.6    & 92.8     & 80.0     & 81.4     & 83.6     &  86.87     &  40.4    &  15.5    &  32.1    &    32.5  &  54.8    &  35.6    & 39.0     &  35.70     &  24.0    \\
\cellcolor[HTML]{EFEFEF}&AutoSTR     & ECCV~\shortcite{zhang2020autostr}       & 94.7     & 90.9     & 94.2     &  81.8    &  81.7    &  -    &   -    &  47.7    &  17.9    & 30.8     &   36.2   & 64.2     & 38.7     &  41.3    & 39.54      &  6.00    \\
\cellcolor[HTML]{EFEFEF}&RoScanner   & ECCV~\shortcite{yue2020robustscanner}       & 95.3 & 88.1 & 94.8 & 77.1 & 79.5 & 90.3 & 87.52 & 43.6 & 7.90  & 41.2 & 42.6 & 44.9 & 46.9 & 39.5 & 38.09 & 48.0 \\
\cellcolor[HTML]{EFEFEF}&PARSeq     & ECCV~\shortcite{BautistaA22PARSeq}       & 97.0 & 93.6 & 97.0 & 86.5 & 88.9 & 92.2 & 92.53 & 63.9 & 16.7 & 52.5 & 54.3 & 68.2 & 55.9 & 56.9 & 52.62 & 23.8 \\
\cellcolor[HTML]{EFEFEF}&LevOCR     & ECCV~\shortcite{levocr}             & 96.6 & 94.4 & 96.7 & 86.5 & 88.8 & 90.6 & 92.27 & 52.8 & 10.7 & 44.8 & 51.9 & 61.3 & 54.0 & 58.1 & 47.66 & 109  \\
\cellcolor[HTML]{EFEFEF}&CornerTF & ECCV~\shortcite{xie2022toward_cornertrans}            & 95.9 & 94.6 & 97.8 & 86.5 & 91.5 & 92.0 & 93.05 & 62.9 & 18.6 & 56.1 & 58.5 & 68.6 & 59.7 & 61.0 & 55.07 & 86.0 \\
\cellcolor[HTML]{EFEFEF}&SIGA       & CVPR~\shortcite{Guan_2023_CVPR_SIGA}     & 96.6 & 95.1 & 97.8 & 86.6 & 90.5 & 93.1 & 93.28 & 59.9 & 22.3 & 49.0 & 50.8 & 66.4 & 58.4 & 56.2 & 51.85 & 113  \\
\cellcolor[HTML]{EFEFEF}&CCD        & ICCV~\shortcite{Guan_2023_ICCV_CCD}      & 97.2 & 94.4 & 97.0 & 87.6 & 91.8 & 93.3 & 93.55 & 66.6 & 24.2 & 63.9 & 64.8 & 74.8 & 62.4 & 64.0 & 60.10 & 52.0 \\
\cellcolor[HTML]{EFEFEF}&LISTER     & ICCV~\shortcite{iccv2023lister}    & 96.9 & 93.8 & 97.9 & 87.5 & 89.6 & 90.6 & 92.72 & 56.5 & 17.2 & 52.8 & 63.5 & 63.2 & 59.6 & 65.4 & 54.05 & 49.9 \\
\cellcolor[HTML]{EFEFEF}&CDistNet & IJCV~\shortcite{zheng2024cdistnet}    & 96.4 & 93.5 & 97.4 & 86.0 & 88.7 & 93.4 & 92.57 & 69.3 & 24.4 & 49.8 & 55.6 & 72.8 & 64.3 & 58.5 & 56.38 & 65.5 \\
\cellcolor[HTML]{EFEFEF}&CAM        & PR~\shortcite{yang2024class_cam}      & 97.4 & 96.1 & 97.2 & 87.8 & 90.6 & 92.4 & 93.58 & 63.1 & 19.4 & 55.4 & 58.5 & 72.7 & 51.4 & 57.4 & 53.99 & 135  \\
\cellcolor[HTML]{EFEFEF}&OTE         & CVPR~\shortcite{Xu_2024_CVPR_OTE}      & 96.4 & 95.5 & 97.4 & 87.2 & 89.6 & 92.4 & 93.08 & -    & -    & -    & -    & -    & -    & -    & -     & 25.2 \\
\cellcolor[HTML]{EFEFEF}&IGTR & TPAMI~\shortcite{du2024igtr}        & \textbf{98.2}     &  95.7    &   \textbf{98.6}   &    \textbf{88.4}  &   \textbf{92.4}   &  95.5    &  94.78                                                                           &  78.4     &    31.9                                                       &      61.3    &   66.5          &   80.2      &    69.3                                                    &  \textbf{67.9}       &    65.07                                                                         &   24.1
                         \\
\cellcolor[HTML]{EFEFEF}&SMTR & AAAI~\shortcite{du2024smtr}  & 97.4 & 94.9 & 97.4 & \textbf{88.4} & 89.9 & 96.2 & 94.02 & 74.2 & 30.6 & 58.5 & 67.6 & 79.6 & \textbf{75.1} & \textbf{67.9} & 64.79  &  15.8 \\
\multicolumn{1}{c|}{\cellcolor[HTML]{EFEFEF}\multirow{-20}{*}{\begin{tabular}[c]{@{}c@{}}A\\R\\M\end{tabular}}} & \multicolumn{2}{l}{\cellcolor[HTML]{D9E1F4}MDiff4STR-B-AR}        &97.9 &	\textbf{96.4} &	98.5 &	87.8 &	91.5 &	96.5 &	94.77 &	\textbf{79.9} &	30.5 &	63.8 &	68.5 &	80.6 &	69.4 &	67.3 &	\textbf{65.72}  & 31.9 \\
\midrule
\cellcolor[HTML]{EFEFEF}&VisionLAN   & ICCV~\shortcite{Wang_2021_visionlan}     & 95.8 & 91.7 & 95.7 & 83.7 & 86.0 & 88.5 & 90.23 & 57.7 & 14.2 & 47.8 & 48.0 & 64.0 & 47.9 & 52.1 & 47.39 & 32.8 \\
\cellcolor[HTML]{EFEFEF}&MGP-STR    & ECCV~\shortcite{mgpstr}       & 96.4 & 94.7 & 97.3 & 87.2 & 91.0 & 90.3 & 92.82 & 55.2 & 14.0 & 52.8 & 48.5 & 65.2 & 48.8 & 59.1 & 49.09 & 148  \\
\cellcolor[HTML]{EFEFEF}&LPV-B      & IJCAI~\shortcite{ijcai2023LPV}       & 97.3 & 94.6 & 97.6 & 87.5 & 90.9 & 94.8 & 93.78 & 68.3 & 21.0 & 59.6 & 65.1 & 76.2 & 63.6 & 62.0 & 59.40 & 35.1 \\
\cellcolor[HTML]{EFEFEF}&CPPD        & TPAMI~\shortcite{du2023cppd}     & 97.6 & 95.5 & 98.2 & 87.9 & 90.9 & 92.7 &  93.80 & 65.5 & 18.6 & 56.0 & 61.9 & 71.0 & 57.5 & 65.8 & 56.63 & 26.8 \\
\multicolumn{1}{c|}{\cellcolor[HTML]{EFEFEF}\multirow{-5}{*}{\begin{tabular}[c]{@{}c@{}}P\\D\\M\end{tabular}}} & \multicolumn{2}{l}{\cellcolor[HTML]{D9E1F4}MDiff4STR-B-PD}        & 97.3 &	94.4 &	97.8 &	87.0 &	90.2 &	96.2 &	93.82 &	77.0 &	29.2 &	60.4 &	65.2 &	77.8 &	52.1 &	65.4 &	61.02 & 31.9 \\
\midrule
\cellcolor[HTML]{EFEFEF}&SRN         & CVPR~\shortcite{yu2020srn}     & 94.8 & 91.5 & 95.5 & 82.7 & 85.1 & 87.8 & 89.57 & 63.4 & 25.3 & 34.1 & 28.7 & 56.5 & 26.7 & 46.3 & 40.14 & 54.7 \\
\cellcolor[HTML]{EFEFEF}&ABINet       & CVPR~\shortcite{fang2021abinet}    & 96.2 & 93.5 & 97.4 & 86.0 & 89.3 & 89.2 & 91.93 & 59.5 & 12.7 & 43.3 & 38.3 & 62.0 & 50.8 & 55.6 & 46.03 & 36.7 \\
\cellcolor[HTML]{EFEFEF}&MATRN       & ECCV~\shortcite{MATRN}   & 96.6 & 95.0 & 97.9 & 86.6 & 90.6 & 93.5 & 93.37 & 63.1 & 13.4 & 43.8 & 41.9 & 66.4 & 53.2 & 57.0 & 48.40 & 44.2 \\
\cellcolor[HTML]{EFEFEF}&BUSNet      & AAAI~\shortcite{Wei_2024_busnet}        & 96.2 & 95.5 & 98.3 & 87.2 & 91.8 & 91.3 & 93.38 & -    & -    & -    & -    & -    & -    & -    & -     & 56.8 \\
\multicolumn{1}{c|}{\cellcolor[HTML]{EFEFEF}\multirow{-5}{*}{\begin{tabular}[c]{@{}c@{}}R\\e\\M\end{tabular}}} & \multicolumn{2}{l}{\cellcolor[HTML]{D9E1F4}MDiff4STR-B-Re}        &\textbf{98.2} &	95.4 &	98.5 &	88.2 &	91.3 &	\textbf{96.9} &	94.73 &	79.0 &	30.2 &	63.8 &	68.3 &	81.2 &	66.9 &	67.1 &	65.21 & 31.9 \\
\midrule
\cellcolor[HTML]{EFEFEF} & \multicolumn{2}{l}{\cellcolor[HTML]{D9E1F4}MDiff4STR-B-LC}        &98.1 &	95.1 &	98.1 &	88.2 &	91.5 &	\textbf{96.9} &	94.63 &	79.3 &	30.2 &	\textbf{64.6} &	67.3 &	80.6 &	64.7 &	67.1 &	64.82 & 31.9 \\
\cellcolor[HTML]{EFEFEF} & \multicolumn{2}{l}{\cellcolor[HTML]{D9E1F4}MDiff4STR-B-BLC}        &98.1 &	95.8 &	98.2 &	88.1 &	91.6 &	\textbf{96.9} &	\textbf{94.81} &	79.8 &	30.2 &	64.0 &	68.2 &	\textbf{80.8} &	66.9 &	67.4 &	65.33  & 31.9 \\
\bottomrule
\end{tabular}}
\caption{Results of MDiff4STR and existing models when trained on synthetic datasets (\textit{ST} + \textit{MJ}) \cite{Synthetic,jaderberg14synthetic}.  represents that the results on \textit{U14M} are evaluated using the model they released.}
\label{tab:syn_sota}
\end{table*}

\section{Bad cases of MDiff4STR}

An inherent limitation of the Masked Denoising Modeling (MDM) framework lies in its inability to remove redundant characters or recover missing ones in the denoising process. As illustrated in Fig.~\ref{fig:badcase}, the first two examples highlight cases where redundant characters persist in the outputs of MDiff4STR, and the denoising process fails to eliminate them. This occurs because MDM is designed to predict masked tokens with deterministic replacements, but lacks the capacity to delete existing tokens. The last two examples in Fig.~\ref{fig:badcase} reveal cases of missing characters in the recognition results. The failure stems from MDM’s inability to identify and insert mask tokens at the appropriate positions during denoising. Since MDM cannot infer where a character is missing without an explicit mask token in place, it lacks the mechanism to correct such omissions, ultimately limiting its capacity to handle insertion operations during inference. These bad cases highlight a future and key enhancement for enabling flexible sequence length adjustment, i.e., insertion and deletion capabilities, within the MDM framework.

\section{Results when trained on synthetic datasets}

Previous STR models typically are trained on synthetic datasets \cite{Synthetic,jaderberg14synthetic} and validated using \textit{Com}. Following this protocol, we also train MDiff4STR on synthetic datasets. In addition to evaluating MDiff4STR on \textit{Com}, we further evaluate our model on the challenging \textit{U14M} dataset to assess its generalization capability. As shown in Tab.~\ref{tab:syn_sota}, MDiff4STR achieves competitive or superior results across multiple benchmarks, despite being trained on less diverse synthetic datasets. All three decoding strategies embedded in MDiff4STR-AR, PD, and Re, consistently outperform previous methods within their respective paradigms. Moreover, our exclusive decoding paradigms LC and BLC yield new state-of-the-art results, surpassing prior bests by 0.03\% and 0.26\% respectively. In summary, MDiff4STR's consistent performance across benchmarks affirms its strong generalization capabilities, making it a practical choice for real-world applications despite limited training data diversity.

\end{document}